\documentclass{article}

 \usepackage[preprint]{neurips_2025}

\usepackage[utf8]{inputenc} 
\usepackage[T1]{fontenc}    
\usepackage{hyperref}       
\usepackage{url}            
\usepackage{booktabs}       
\usepackage{amsfonts}       
\usepackage{nicefrac}       
\usepackage{microtype}      
\usepackage{xcolor}         
\usepackage{amsmath, amssymb} 
\usepackage{graphicx} 
\usepackage{caption}
\usepackage{subcaption}

\title{Linearized Optimal Transport for Analysis of High-Dimensional Point-Cloud and Single-Cell Data}

\author{%
  Tianxiang Wang\\
  University of California, San Diego\\
  \texttt{tiw014@ucsd.edu} \\
  \And
  Yingtong Ke\\
  University of California, San Diego\\
  \texttt{y2ke@ucsd.edu} \\
  \And
  Dhananjay Bhaskar\\
  University of Wisconsin-Madison\\
  \texttt{dhananjay.bhaskar@wisc.edu} \\
  \And
  Smita Krishnaswamy\\
  Yale University \\
  \texttt{smita.krishnaswamy@yale.edu} \\
  \And
  Alexander Cloninger\\
  University of California, San Diego\\
  \texttt{acloninger@ucsd.edu} \\
}

\begin{document}

\maketitle

\begin{abstract}
  Single-cell technologies generate high-dimensional point clouds of cells, enabling detailed characterization of complex patient states and treatment responses. Yet each patient is represented by an irregular point cloud rather than a simple vector, making it difficult to directly quantify and compare biological differences between individuals. Nonlinear methods such as kernels and neural networks achieve predictive accuracy but act as black boxes, offering little biological interpretability.

    To address these limitations, we adapt the Linear Optimal Transport (LOT) framework to this setting, embedding irregular point clouds into a fixed-dimensional Euclidean space while preserving distributional structure. This embedding provides a principled linear representation that preserves optimal transport geometry while enabling downstream analysis. It also forms a registration between any two patients, enabling direct comparison of their cellular distributions. Within this space, LOT enables: (i) \textbf{accurate and interpretable classification} of COVID-19 patient states, where classifier weights map back to specific markers and spatial regions driving predictions; and (ii) \textbf{synthetic data generation} for patient-derived organoids, exploiting the linearity of the LOT embedding. LOT barycenters yield averaged cellular profiles representing combined conditions or samples, supporting drug interaction testing.

    Together, these results establish LOT as a unified framework that bridges predictive performance, interpretability, and generative modeling. By transforming heterogeneous point clouds into structured embeddings directly traceable to the original data, LOT opens new opportunities for understanding immune variation and treatment effects in high-dimensional biological systems.
\end{abstract}

\section{Introduction}
Modern single-cell technologies enable detailed profiling of millions of individual cells, each represented as a high-dimensional point in marker space. However, a fundamental challenge remains: cells are not directly aligned across individuals. Each patient yields a unique, unordered point cloud of cells, making population-level comparisons and model training difficult.

Optimal Transport (OT) has emerged as a principled way to compare such single-cell distributions by computing the cost of transforming one point cloud into another. Prior studies have successfully applied OT, including Geodesic Sinkhorn \citep{huguet2023geodesic} and PhEMD \citep{chen2020phemd}, as well as OT-inspired approaches such as Trellis \citep{ramos2023trellis}, to quantify differences between cell populations, especially in the absence of cell-level correspondence. However, traditional OT returns only a distance between two distributions, which is useful for quantifying divergence but limited in scope. The transport plan underlying this distance is pair-specific and its size scales with the number of cells in both samples, making it difficult to interpret biologically or to reuse across comparisons. Therefore, even while OT distances are useful for measuring dissimilarity, they do not provide a consistent representation of each sample that can be directly leveraged for learning, interpretability, or generative modeling. In contrast, LOT compares each distribution to a fixed reference, producing a fixed-dimensional embedding that places all samples in a common Euclidean space, enabling downstream analysis.

To address these limitations, we adopt the Linear Optimal Transport (LOT) framework \citep{moosmuller2020linear}, which embeds point clouds into a shared, Euclidean representation space using transport maps to a fixed reference distribution. This formulation preserves distributional geometry, enables differentiable model learning, and yields interpretable vector representations for each sample.

This work demonstrates two core capabilities of LOT embeddings:
\begin{itemize}
    \item \textbf{Interpretable classification} — LOT embeds each distribution into a fixed-length vector in a Euclidean space defined relative to a common reference distribution. In this embedding, coordinates are defined with respect to the set of reference points and capture transport displacements along marker directions, integrating contributions from many cells. The result is a consistent coordinate system shared across all samples, so linear models can be trained directly in this space. Classifier weights can then be mapped back onto marker axes and reference locations, enabling interpretation of decision boundaries in terms of biological features such as shifts in marker expression or enrichment of specific cell populations.

    \item \textbf{Synthetic data generation} — LOT embeds each distribution into a Euclidean vector space with well-defined linear structure and a trivial pre-image. Unlike nonlinear embeddings, where the geometry of the space is complex and difficult to interpret, LOT provides clear semantics: vectors can be added, averaged, or interpolated in embedding space, and the results can always be mapped back to marker features. This enables the construction of synthetic embeddings that approximate intermediate or mixed biological states, a capability that is otherwise intractable in raw single-cell data where samples differ in size and lack canonical alignment.

\end{itemize}

We illustrate these capabilities using two complementary datasets, one from COVID-19 immune profiling \citep{kuchroo2022multiscale} and another from patient-derived cancer organoids \citep{ramos2023trellis}, showing that LOT supports both supervised and unsupervised discovery in high-dimensional immune data. 

Across both settings, LOT provides a unified framework to interpret variability, enabling the classification of patient states, the generation and testing of synthetic embeddings, and the identification of dominant response patterns shared across individuals.  

\subsection{Contributions}
\begin{enumerate}
    \item \textbf{Applying LOT embeddings to single-cell data:} While optimal transport is a principled way to compare distributions, it is inherently defined pairwise and has seen limited use in large-scale single-cell studies. Here,  building on the LOT framework \citep{moosmuller2020linear}, we demonstrate that LOT embeddings provide a fixed-dimensional representation of each patient’s immune point cloud relative to a common reference, enabling rigorous cross-patient and cross-condition comparison. This application establishes LOT as a practical tool for making heterogeneous single-cell datasets directly comparable in biomedical settings.

    \item \textbf{Predictive and interpretable modeling:} Our framework achieves strong classification of COVID-19 patient states, but more importantly, it does so while remaining fully traceable to the original point-cloud data. Unlike typical nonlinear methods (e.g., kernels, neural networks) that provide accuracy without transparency, LOT embeddings preserve a direct link between the classifier’s decision boundary and the underlying point-cloud structure. This means that predictive weights can be projected back onto specific markers and spatial regions, revealing which biological signals dominate the classification. As a result, our approach not only matches the predictive power of cutting-edge techniques, but it also converts such predictions into interpretable scientific insight that can be understood and is based on the original data. 
    
    \item \textbf{Synthetic data generation from non-uniform distributions:} In many analyses, creating synthetic datasets is helpful, for example by combining, averaging, or interpolating samples to explore hypotheses or assess additive versus interactive effects. Raw point clouds, however, differ in size and geometry, making such operations challenging to define in a rigorous way. By embedding into the common LOT space, we enable linear operations such as addition, averaging, and interpolation to be applied meaningfully. This opens a powerful new avenue for creating and exploring synthetic embeddings, extending analysis to operations on non-uniform distributions that are not achievable in their raw form.
\end{enumerate}

\section{Background}
\subsection{Optimal Transport (OT)}
Optimal Transport (OT) provides a rigorous way to compare probability distributions by computing the minimal cost of transforming one distribution into another. The associated Wasserstein distances \(W_p\) form a family of metrics parameterized by \(p\). In particular, \(W_1\) (the Earth Mover’s distance) measures average transport cost, while \(W_2\) emphasizes squared Euclidean distances and induces a Riemannian geometry on probability measures, supporting concepts such as geodesics and barycenters.  With advances like the entropic regularization \citep{cuturi2013sinkhorn}, OT has become computationally practical and widely adopted across scientific and machine learning domains.

\subsection{Linear Optimal Transport (LOT)}
Linear Optimal Transport (LOT) was developed to address the limitations of classical OT, which requires computing distances between all pairs of samples. Instead of pairwise comparison, each distribution $\mu$ is transported to a fixed reference distribution $\sigma$, and the resulting displacement is linearized into a finite-dimensional vector \citep{wang2013linear}. This construction provides a Euclidean embedding that linearizes Wasserstein geometry around the reference.

Formally, LOT defines an embedding
\[
F_\sigma : \mathcal{P}(\mathbb{R}^d) \to L^2(\mathbb{R}^d, \sigma), 
\quad \mu \mapsto T_\sigma^\mu,
\]
where $T_\sigma^\mu$ is the optimal transport map from the reference $\sigma$ to the target distribution $\mu$, and $L^2(\mathbb{R}^d,\sigma)$ is the Hilbert space of square-integrable functions with respect to $\sigma$ \citep{moosmuller2020linear}. In practice, we discretize $\sigma$ by choosing $m$ reference points and compute the displacement field
\[
u(x) = T_\sigma^\mu(x) - x, \quad x \sim \sigma.
\]
Flattening these displacements across all reference points yields the LOT embedding vector
\[
z = \mathrm{vec}\!\big(u(x_1), u(x_2), \ldots, u(x_m)\big),
\]
which transforms irregular point clouds into fixed-length vectors in Euclidean space.

These embeddings make downstream learning feasible: linear classifiers, regression models, and clustering methods can all be applied directly in LOT space. Crucially, Moosmüller and Cloninger \citep{moosmuller2020linear} established that under specific assumptions on the underlying distribution classes (e.g., rigid transformations or certain perturbations), linear classifiers in LOT space can achieve perfect classification accuracy. Beyond classification, LOT also supports synthesis, because its linear structure allows averaging or interpolating embeddings in Euclidean space, which corresponds to meaningful synthetic distributions when mapped back. This property is not well defined on raw point clouds of variable size. In addition, LOT reduces computational cost because it requires only one OT computation per sample rather than all pairwise distances.

Originally introduced in image analysis, where it enabled quantification of variation in microscopy \citep{kolouri2016continuous} and facial datasets \citep{wang2013linear}, LOT has since been extended to point set classification \citep{rabbi2024lotclassification}, scientific domains such as particle physics \citep{cai2020lotcollider}, supervised learning of sheared distributions \citep{khurana2023sheared}, dimensionality reduction with provable guarantees \citep{cloninger2025lotdr}, and software frameworks for point cloud analysis \citep{linwu2025pylot}. In each case, the appeal of LOT lies in turning irregular, high-dimensional distributions into structured vector representations that can be analyzed with standard machine learning tools.

\subsection{Gap in the Literature}

Despite the broad adoption of OT and LOT in imaging, vision, NLP, and physics, their use in single-cell biology remains limited. Most single-cell OT methods rely on pairwise distances or trajectory alignments, rather than providing a shared vector space that enables both interpretable classification and straightforward generation of synthetic embeddings. Our work addresses this gap by adapting LOT to single-cell immune and patient-derived cancer organoid datasets, demonstrating how it can support classification of patient states, construction of synthetic conditions, and interpretation of embedding directions in terms of biologically meaningful markers and responses.

\section{Using LOT for Interpretable Classification}
\subsection{Methods}

Our analytical pipeline proceeds in three stages: (i) embed each patient’s single-cell point cloud into a common Euclidean space using Linear Optimal Transport (LOT), (ii) train a linear Support Vector Machine (SVM) to classify embeddings by clinical state, and (iii) interpret the resulting classifier weights through spectral co-clustering, which reveals coupled groups of markers and spatial reference points. This chain of algorithms leverages LOT for comparability, SVMs for predictive discrimination, and co-clustering for biological interpretability. All analyses were run on a standard CPU workstation; no GPUs or clusters were required.

\subsubsection{LOT embedding}
To enable classification from single-cell point clouds, we applied the Linear Optimal Transport (LOT) framework. Each sample distribution $\mu_i$ was mapped to a shared reference distribution $\sigma$ via the barycentric transport map $T_\sigma^{\mu_i}$. The displacement fields were evaluated at $m$ reference points and vectorized into embeddings $z_i \in \mathbb{R}^p$, with $p = m \times d$ and $d$ denoting the marker dimension. This produced fixed-length vectors in a common Euclidean coordinate system, ensuring that all samples could be compared directly and analyzed using linear classifiers and spectral co-clustering.

\subsubsection{SVM classification}
In the second stage, we trained linear SVMs on the LOT embeddings $z_i$ to classify patients by clinical state (e.g., healthy vs.\ COVID-positive). In the binary case, SVM optimization is formulated with labels
\[
y_i \in \{-1, +1\},
\]
leading to the decision function
\[
f(z) = w^\top z + b,
\]
where $w \in \mathbb{R}^p$ is the learned weight vector and $b$ is the bias. Crucially, because $w$ has the same dimension as the LOT embedding, it can be reshaped back to the reference grid of points and features, enabling spatially grounded biological interpretation.

\subsubsection{Spectral co-clustering}
Finally, to interpret the classifier weights, we reshape $w$ into a matrix
\[
W \in \mathbb{R}^{m \times d},
\]
where rows index reference points and columns index markers. Spectral co-clustering is then applied to $W$, treating it as a bipartite graph between spatial regions and proteins. This yields biclusters $(C_r^k, C_c^\ell)$ that highlight coherent marker–location signatures associated with immune state differences, making the full pipeline both predictive and interpretable.
\paragraph{Notation.}
Let $K$ and $L$ denote the numbers of row (reference) and column (marker) clusters, respectively.
We index row clusters by $k=1,\dots,K$ and column clusters by $\ell=1,\dots,L$.
Let $C_r^{(k)} \subset \{1,\dots,m\}$ be the set of reference indices in row cluster $k$ and 
$C_c^{(\ell)} \subset \{1,\dots,d\}$ the set of marker indices in column cluster $\ell$.
A bicluster is the pair $(C_r^{(k)}, C_c^{(\ell)})$.

\subsection{COVID Data Description}
We analyzed single-cell immune profiling data from 150 individuals, including COVID-19-positive patients and healthy hospital workers (HCWs). For each patient $i$, the data form a point cloud of immune cells in $\mathbb{R}^{16}$; The 16 features are: CD14, CD183/CXCR3, CD196/CCR6, CD3, CD4, CD8, FSC, Fixable Aqua, Granzyme B, IFN-$\gamma$, IL-2, IL-6, IL-17A, IL-4, SSC, and TNF-$\alpha$. Labels were binarized by mapping HCW to \emph{healthy} (0) and all patient labels $\{0.0,1.0\}$ to \emph{sick} (1); two subjects without usable labels were excluded (final $N=148$).

\subsection{Results}

\subsubsection{Classification Performance with LOT Embeddings}
We first evaluated the ability of LOT embeddings to support accurate patient classification. 
A linear SVM was trained using an 80/20 stratified train--test split. 
The model achieved a test accuracy of \textbf{90.0\%}, confirming strong separation between COVID-positive and healthy individuals in LOT space.  

\paragraph{Confusion Matrix and Metrics.}
Performance metrics are summarized in Table~\ref{tab:svm_metrics}. 
Precision and recall values were balanced across both groups, with COVID-positive patients classified with 0.94 precision and 0.89 recall, and healthy controls classified with 0.83 precision and 0.91 recall.  

\begin{table}[ht]
\centering
\caption{Linear SVM classification performance on LOT embeddings.}
\label{tab:svm_metrics}
\begin{tabular}{lccc}
\hline
Class & Precision & Recall & F1-score \\
\hline
COVID-positive & 0.94 & 0.89 & 0.91 \\
Healthy        & 0.83 & 0.91 & 0.87 \\
\hline
\textbf{Overall Accuracy} & \multicolumn{3}{c}{0.90} \\
\textbf{AUC} & \multicolumn{3}{c}{0.95} \\
\hline
\end{tabular}
\end{table}

\paragraph{ROC Analysis.}
The ROC curve (AUC = 0.95) illustrates the high discriminative power of the decision boundary.

\subsubsection{Interpretation of Co-clustered Signatures}
To investigate the interpretability of the learned classifier, 
we analyzed the SVM weight matrix after spectral co-clustering. 
Rows (LOT reference points) and columns (biological markers) were reordered by their assigned clusters, 
yielding a structured heatmap where red values indicate strong positive association with COVID-positive prediction 
and blue values indicate strong negative (healthy-associated) contributions.  

Coherent red/blue blocks in this visualization highlight spatial--feature combinations with aligned discriminative influence. 
To summarize these effects, we computed the \textbf{mean absolute weight matrix} across the 7 row clusters and 5 feature clusters, 
producing a $7 \times 5$ grid of bicluster importance values. 
Brighter values indicate stronger overall influence, independent of prediction direction. 
Clusters 1 and 5 exhibited the highest mean absolute weights and were therefore prioritized for downstream biological interpretation.  

\begin{figure}[ht]
    \centering
    \includegraphics[width=0.8\linewidth]{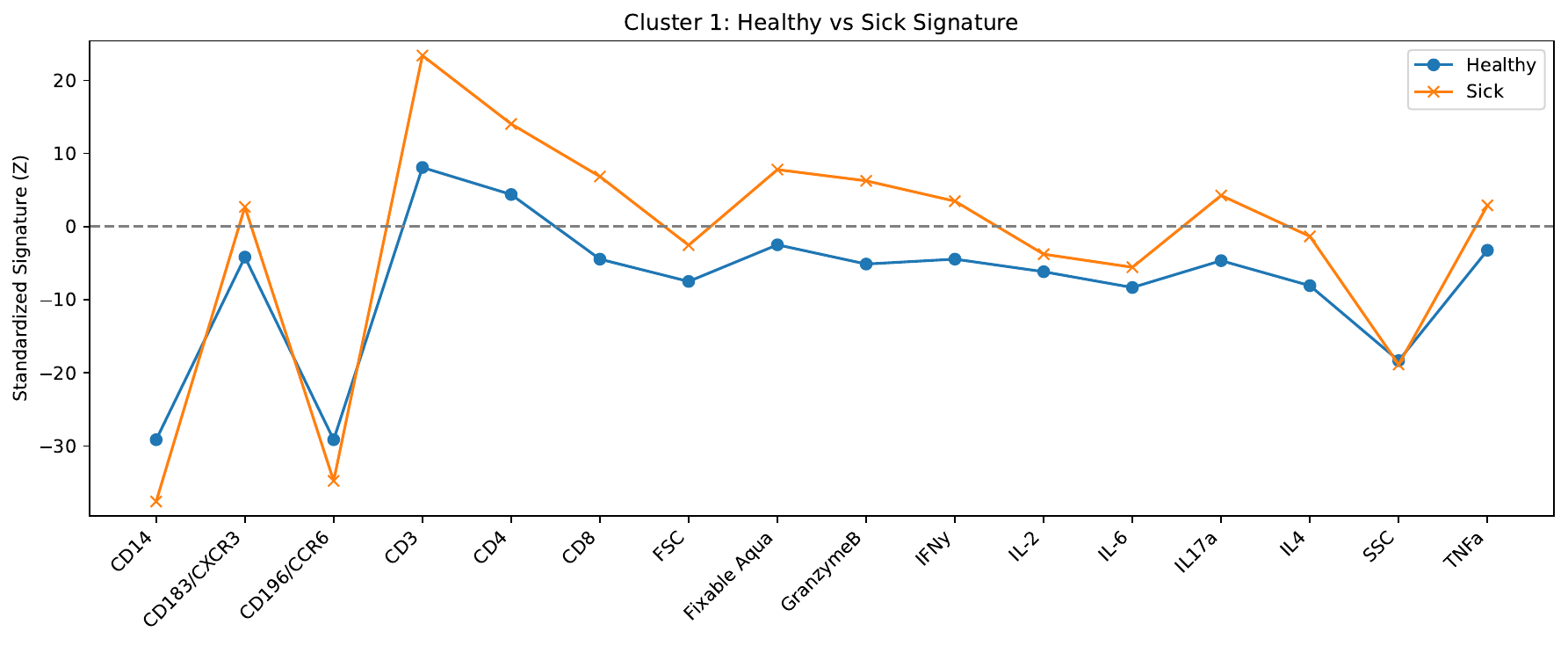}
    \vspace{0.5em}
    \includegraphics[width=0.8\linewidth]{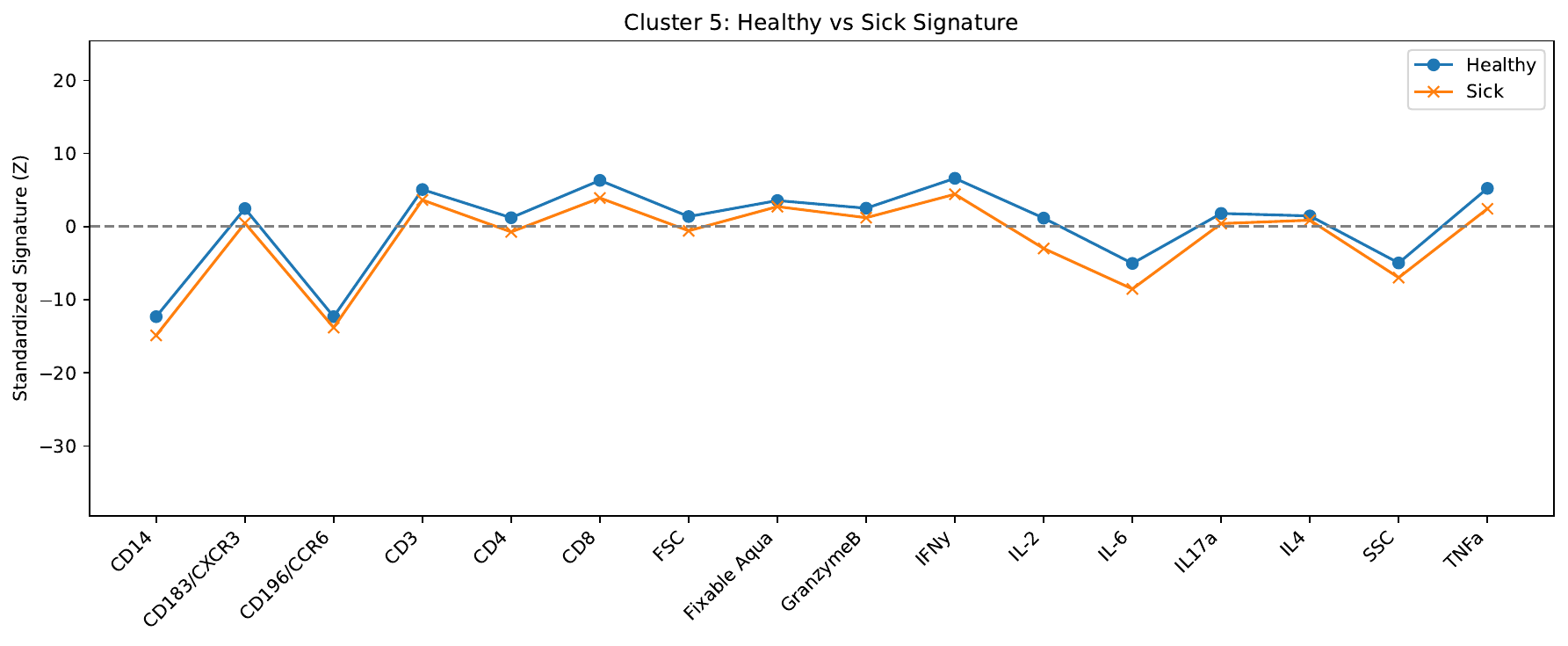}
    \caption{Cluster activation signatures contrasting COVID-positive and healthy groups. Each line plot shows standardized marker-level activations ($Z$-scores) for one co-cluster. \textbf{Top:} Cluster~1 shows broad up-regulation of T-cell lineage and inflammatory markers in COVID-positive patients, consistent with heightened immune activation. \textbf{Bottom:} Cluster~5 shows highly similar marker profiles between groups, but with consistently lower levels in COVID-positive patients, supporting its interpretation as a resting or naïve cluster. Together, these signatures highlight how distinct spatial-feature clusters capture biologically interpretable differences in immune response between conditions.}
    \label{fig:covid_signatures}
\end{figure}

\paragraph{Cluster Signature Analysis.}
To connect these discriminative biclusters with patient-level biology, 
we performed a cluster activation analysis:  

\begin{enumerate}
    \item \textbf{Cluster reference extraction.} 
    For each of the 7 row clusters, we isolated the corresponding LOT reference points from each patient’s embedding.  
    \item \textbf{Group-wise pooling.} 
    Patients were split into COVID-positive and healthy groups. 
    Within each group, we pooled all LOT values from the cluster and computed the mean activation vector (16-dimensional).  
    \item \textbf{Global normalization.} 
    We computed the global mean $\mu_{\text{global}}$ and standard deviation $\sigma_{\text{global}}$ 
    across all patients and all LOT values. 
    Each group’s cluster-level mean vector $x_{\text{group}}$ was then standardized using
    \[
    \mathbf{Z}_{\text{group}} \;=\; 
    \frac{\mathbf{x}_{\text{group}} - \boldsymbol{\mu}_{\text{global}}}
     {\boldsymbol{\sigma}_{\text{global}} / \sqrt{n_{\text{cells}}}},
    \]
    where $x_{\text{group}}$ is the group-level mean vector, 
    $\mu_{\text{global}}$ and $\sigma_{\text{global}}$ are the global mean and standard deviation, 
    and $n_{\text{cells}}$ is the total number of LOT values in the group for that cluster.
  
    \item \textbf{Signature plotting \& validation} 
    These standardized vectors were visualized as ``cluster signatures''(Figure~\ref{fig:covid_signatures}),
    where each curve represents marker-level activation patterns within a spatial cluster. Cluster~1 revealed a marked difference between COVID-positive and healthy patients, with the former showing a pronounced activated phenotype and broadly elevated signals. This interpretation is supported by KS tests, which showed highly significant enrichment across nearly all activation-associated markers (FDR $<0.001$; Table~\ref{tab:ks_clusters}).
    By contrast, Cluster~5 showed a more modest distinction between groups, with flatter profiles consistent with resting or na\"{\i}ve cell populations. Here, only a limited subset of cytokines (e.g., IFN-$\gamma$, TNF-$\alpha$, IL-2) and lineage markers (CD3, CD8) reached significance, generally in the negative direction (FDR $<0.05$; Table~\ref{tab:ks_clusters}).

    \begin{table}[ht]
    \centering
    \caption{KS test results for Cluster~1 (activated) and Cluster~5 (resting/na\"{\i}ve). Signs indicate direction of COVID-positive enrichment. Correction method: FDR (Benjamini–Hochberg)}
    \label{tab:ks_clusters}
    \setlength{\tabcolsep}{2pt}
    \begin{minipage}{0.50\linewidth}
    \centering
    \scriptsize
    \textbf{Cluster 1 (Activated)} \\
    \begin{tabular}{lccc}
    \hline
    Feature & KS\_p & KS\_FDR & Sign \\
    \hline
    IL17a & 0.0000 & 0.0000 & + \\
    GranB & 0.0000 & 0.0000 & + \\
    SSC & 0.0000 & 0.0000 & + \\
    IL4 & 0.0000 & 0.0000 & + \\
    CD8 & 0.0000 & 0.0000 & + \\
    IL-6 & 0.0000 & 0.0000 & + \\
    CXCR3 & 0.0000 & 0.0000 & + \\
    CD3 & 0.0000 & 0.0000 & + \\
    IFN-$\gamma$ & 0.0000 & 0.0000 & + \\
    IL-2 & 0.0001 & 0.0002 & + \\
    CD4 & 0.0003 & 0.0003 & + \\
    TNF-$\alpha$ & 0.0006 & 0.0007 & + \\
    \hline
    \end{tabular}
    \end{minipage}\hfill
    \begin{minipage}{0.50\linewidth}
    \centering
    \scriptsize
    \textbf{Cluster 5 (Resting)} \\
    \begin{tabular}{lccc}
    \hline
    Feature & KS\_p & KS\_FDR & Sign \\
    \hline
    IFN-$\gamma$ & 0.0003 & 0.0050 & $-$ \\
    TNF-$\alpha$ & 0.0021 & 0.0157 & $-$ \\
    IL-2 & 0.0033 & 0.0157 & $-$ \\
    FSC & 0.0039 & 0.0157 & $-$ \\
    IL-6 & 0.0066 & 0.0211 & $-$ \\
    CD8 & 0.0186 & 0.0497 & $-$ \\
    CD3 & 0.0217 & 0.0497 & $-$ \\
    \hline
    \end{tabular}
    \end{minipage}

    \vspace{0.5em} 
    \scriptsize \textit{Note: “+” = higher in COVID-positive patients, “–” = lower.}
    \end{table}
    
\end{enumerate}

This analysis revealed distinct activation patterns between COVID-positive and healthy groups, 
with several clusters showing consistent marker-level differences. 
These co-cluster signatures provide biologically interpretable evidence that the LOT embedding 
captures meaningful immune variation across patient conditions.

\section{Using LOT for Data Generation and Treatment Effectiveness}
\subsection{Methods}
Our pipeline for analyzing treatment effects proceeds in three main steps: 
first, patient–culture–replicate samples are embedded into a common LOT space; second, block-wise normalization with internal controls (DMSO) centers each block to its baseline; and third, contrasts between additive and combined treatments are summarized by singular value decomposition (SVD) and further interpreted with spectral biclustering. This chain of algorithms leverages LOT to make heterogeneous patient-derived organoid data directly comparable, while providing interpretable axes of treatment-induced variation. All analyses were run on a standard CPU workstation; no GPUs or clusters were required.

\subsubsection{Block-Wise Normalization}
To correct for baseline and batch effects, we performed within-block normalization using a designated control condition. 
For each patient–culture–replicate block containing a DMSO control, the mean LOT embedding of the control was subtracted from all other conditions within the same block. Control samples were subsequently excluded from downstream analyses, and blocks without a control were left unchanged. This step ensures that treatment effects are interpreted relative to a block-specific baseline. Importantly, this centering is only possible because LOT embeddings provide a fixed coordinate system across patients and conditions, which is not available in raw point cloud space.

\subsubsection{Construction of the Contrast Matrix $\Delta$}
To quantify potential interaction effects between different treatment conditions, we constructed a contrast matrix $\Delta$ that measures the deviation of a combined condition from the additive expectation of its individual components. Formally, for any triplet of matched conditions consisting of two single treatments and their combination, 
we define

\begin{equation}
\Delta = \text{LOT}(\text{combined})
- \tfrac{1}{2}\big( \text{LOT}(\text{treatment 1})
+ \text{LOT}(\text{treatment 2}) \big)
\end{equation}

Each row of $\Delta$ corresponds to one matched triplet, represented as a vector in $\mathbb{R}^p$ with $p$ equal to the LOT embedding dimension.  
Stacking across all triplets yields the matrix $\Delta \in \mathbb{R}^{n \times p}$, where $n$ is the number of valid triplets included in the analysis.

\subsubsection{Singular Value Decomposition of $\Delta$}
To identify dominant patterns in the contrast matrix $\Delta$, we applied Singular Value Decomposition (SVD):
\[
\Delta = U \Sigma V^\top,
\]
where $U \in \mathbb{R}^{n \times r}$ contains the left singular vectors, 
$\Sigma \in \mathbb{R}^{r \times r}$ is diagonal with singular values, 
and $V \in \mathbb{R}^{p \times r}$ contains principal directions in LOT space, with $r=\min(n,p)$.
Because $\Delta$ is extremely high-dimensional, we assessed whether large singular values correspond to structured biological signal or random noise by comparing the empirical spectrum to \textbf{Mar\v{c}enko--Pastur (MP) theory}.  

MP theory provides a null model for the eigenvalue distribution of large random matrices whose entries are independent and identically distributed. It predicts that, under the null, the eigenvalues of $\tfrac{1}{n}\,\Delta\Delta^\top$ converge to a deterministic distribution supported on
\[
\lambda \in \Big[ \sigma^2 (1 - \sqrt{\gamma})^2, \, \sigma^2 (1 + \sqrt{\gamma})^2 \Big],
\]
where $\sigma^2$ is the entrywise variance and $\gamma = \tfrac{p}{n}$ is the aspect ratio.
\textbf{In our experiments, we analyze the spectrum of the \emph{column-centered} contrasts $D=\Delta-\bar{\Delta}$ (subtracting the mean across triplets for each feature) and use the unbiased covariance scaling, so the plotted eigenvalues satisfy $\lambda_i = s_i(D)^2/(n-1) = s_i(D)^2/(B-1)$.}
Any eigenvalue lying significantly above this bulk interval is interpreted as evidence for a low-rank, structured signal embedded in high-dimensional noise (equivalently, singular values $s_i(D)$ with $s_i(D)^2/(n-1)$ above the MP upper edge). To facilitate interpretation, each \emph{un-centered} contrast vector is projected into the top components of $V$ (coordinates $\Delta V = U\Sigma$; we refer to $U\Sigma$ as scores). The projection onto the first two singular directions defines a $\Delta$-SVD space in which each patient–replicate is represented as a low-dimensional point, preserving the dominant axes of deviation from additivity.

\subsubsection{Spectral Biclustering}

We apply spectral \emph{biclustering} to the $\Delta$ matrix to identify subsets of reference points and markers that jointly drive treatment-specific deviations. 
For the algorithmic details of the spectral approach, see Section~3 (co-clustering); here we use the biclustering variant to accommodate signed values in $\Delta$.

\subsection{PDO Data Description}
\begin{figure}[ht]
    \centering
    \begin{minipage}[t]{0.48\columnwidth}
        \centering
        \includegraphics[width=\linewidth]{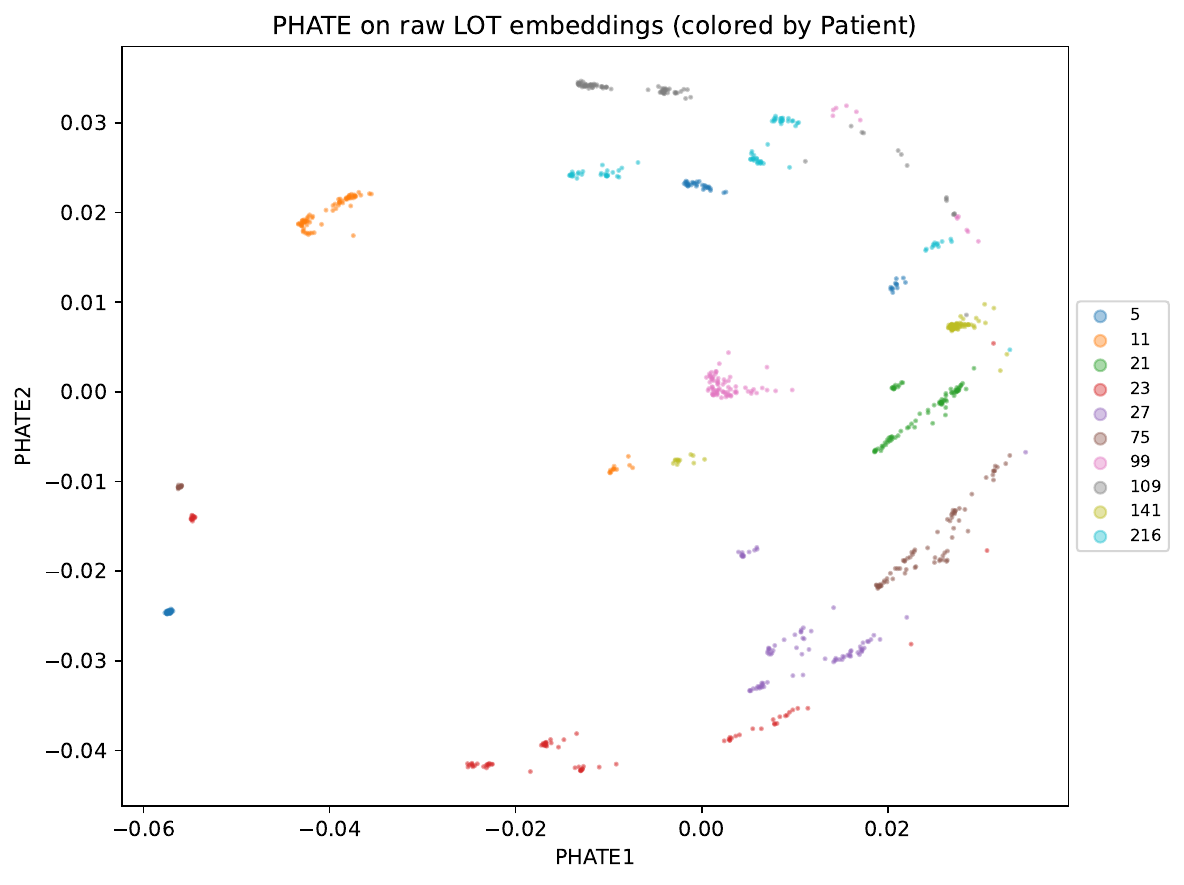}
    \end{minipage}\hfill
    \begin{minipage}[t]{0.48\columnwidth}
        \centering
        \includegraphics[width=\linewidth]{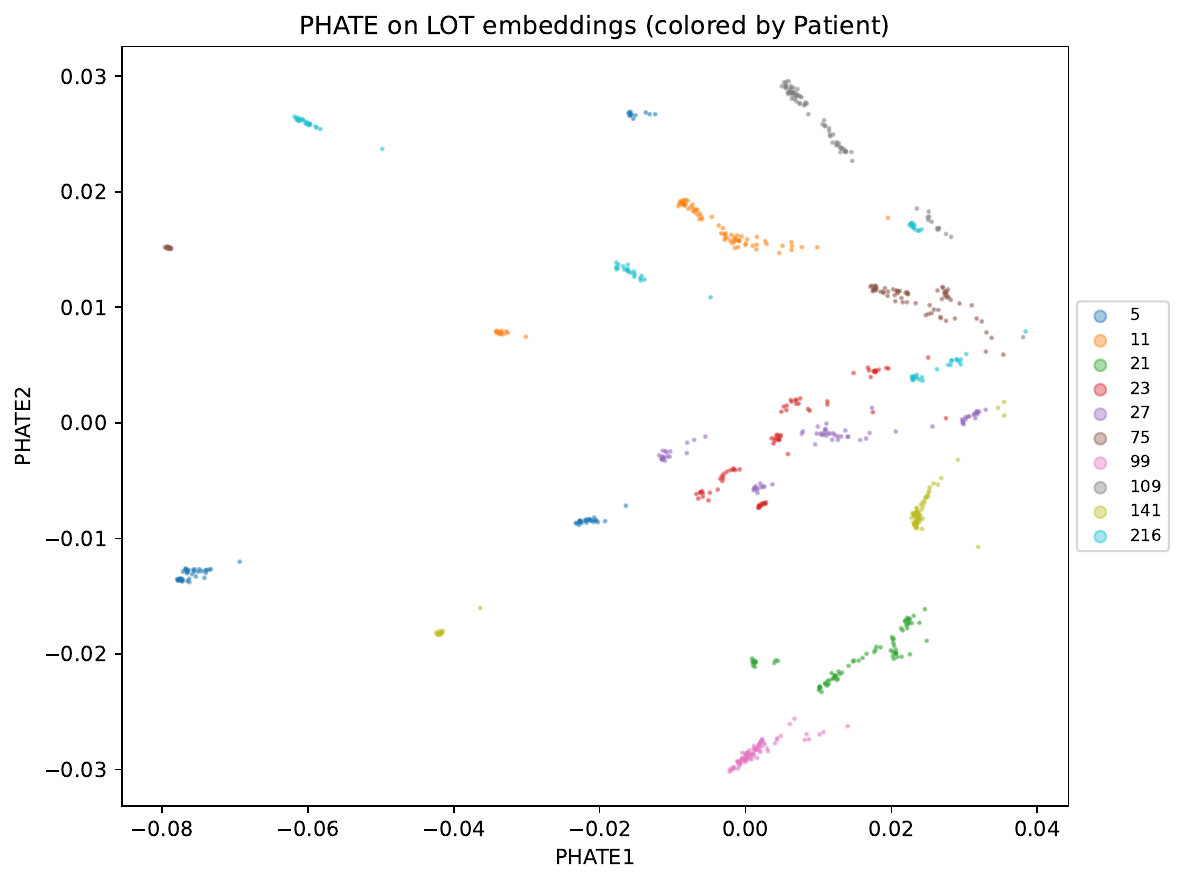}
    \end{minipage}
    \caption{PHATE on \emph{sample-level} LOT embeddings, colored by patient. \textbf{Left:} raw LOT embeddings. \textbf{Right:} after per–patient DMSO control subtraction.}
    \label{fig:phate_pdo_patient}
\end{figure}

To investigate treatment-specific immune profiles in patient-derived organoids (PDOs), we analyzed a high-dimensional single-cell proteomics dataset consisting of over two million individual cell measurements. Each cell was profiled with 44 protein markers, yielding a 44-dimensional feature vector, and annotated with metadata including patient ID, treatment label, replicate ID, and batch information. Treatments included a diverse set of experimental conditions, notably including a single treatment condition (C), another single treatment (F), and a combined treatment (CF), among others. Prior to analysis, we removed ambiguous or duplicate replicates (labeled \texttt{AA}, \texttt{BB}, and \texttt{CC}), resulting in a filtered dataset of $1\,922\,204$ cells. These cells were grouped by unique \texttt{(Patient, Culture, Replicate, Treatment)} combinations to form structured point clouds, where each point represents one immune cell in $\mathbb{R}^{44}$. We identified 840 valid, non-empty point clouds across all patients and treatment conditions. To visualize overall structure, we ran PHATE \citep{moon2019phate} on the \emph{sample-level} LOT embeddings (Fig.~\ref{fig:phate_pdo_patient}). Each point is one Patient–Culture–Replicate–Treatment embedding, colored by patient. We also show the effect of per–patient DMSO control subtraction (for each patient, subtracting that patient’s mean DMSO embedding from all of their embeddings), which LOT enables as a simple linear operation.

\subsection{Results}
\subsubsection{Singular Value Spectrum of $\Delta$}
To evaluate whether the immune response under the combined treatment (CF) could be explained as a simple additive combination of the single treatments (C and F), we constructed the contrast matrix $\Delta$ as described in the Methods. Each row of $\Delta$ represents the deviation of a CF sample from the additive expectation $\tfrac{1}{2}(\text{LOT(C)} + \text{LOT(F)})$, highlighting non-additive, interaction-driven treatment effects at the level of LOT embeddings.

\begin{figure}[t]
    \centering
    \includegraphics[width=0.7\columnwidth]{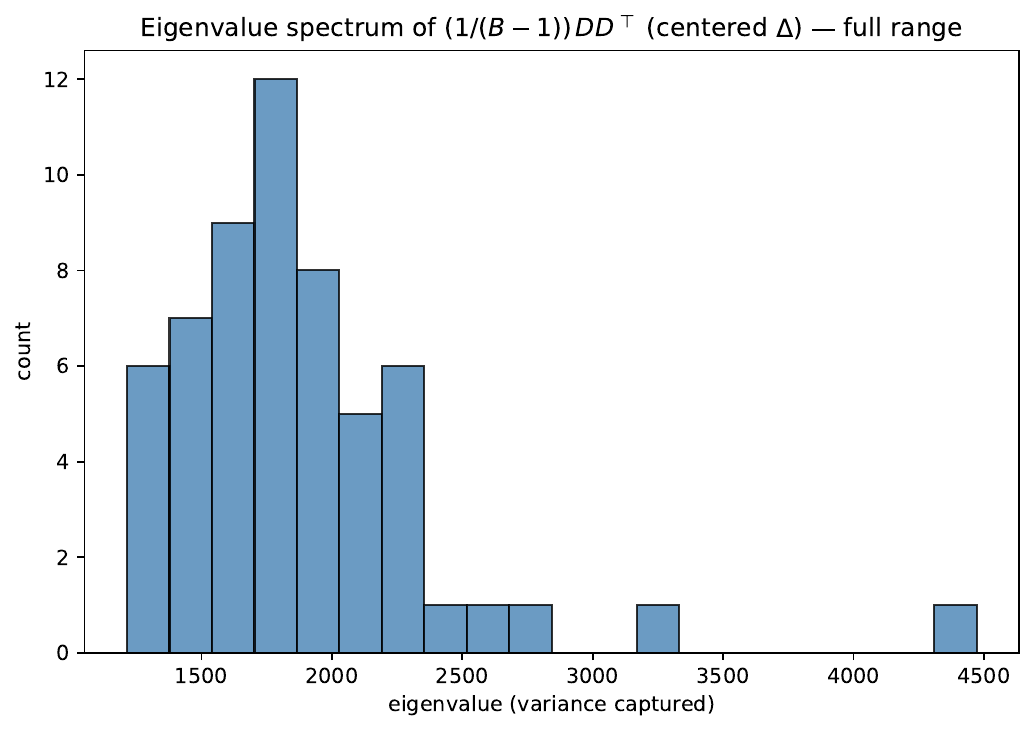}
    \caption{Eigenvalue spectrum of $(1/(B-1))\,(\Delta - \bar{\Delta})(\Delta - \bar{\Delta})^\top$ (\emph{column-centered}; $B=\langle\text{value}\rangle$). The bulk follows the Mar\v{c}enko--Pastur distribution, with two dominant outliers indicating a structured immune response pattern.}
    \label{fig:eigen_spectrum}
\end{figure}

We then examined both (i) the eigenvalue spectrum of $(1/(B-1))\,D D^\top$ (with $D=\Delta-\bar{\Delta}$, \emph{column-centered}) against the Mar\v{c}enko--Pastur (MP) law, and (ii) the raw singular values of $\Delta$. The empirical eigenvalue spectrum (Fig.~\ref{fig:eigen_spectrum})showed a bulk of values concentrated within a narrow range ($\lambda \approx 2200$–$2400$), together with two clear outliers at the top ($\lambda_{1} \approx 4471.6$, $\lambda_{2} \approx 3174.7$). For reference, the top raw singular values of $\Delta$ are $s_1=516.59$, $s_2=430.21$, … (units of the LOT embedding).
\emph{Note that the MP plot is computed from the column-centered matrix $D=\Delta-\bar{\Delta}$; its eigenvalues satisfy $\lambda_i = s_i(D)^2/(B-1)$.} The strong separation of these two components from the MP bulk indicates that they capture structured, low-rank treatment signal rather than random variation. This interpretation is supported by Mar\v{c}enko--Pastur (MP) theory, which predicts that under a null model of independent, identically distributed noise, eigenvalues should remain confined to a narrow bulk interval determined by the matrix aspect ratio.  The emergence of two eigenvalues far above this theoretical bulk confirms the presence of meaningful, low-dimensional structure in the treatment response captured by $\Delta$.

In our case, two outlier singular values indicate the existence of a coherent immune response pattern to the combined treatment (CF) that is consistently expressed across patients.  
The top right-singular vector of $\Delta$ defines the principal LOT-space direction 
along which CF deviates from the additive expectation of C and F. This direction captures the dominant mode of treatment interaction and forms the basis for downstream interpretation of marker-level signatures.

\begin{figure}[ht] 
    \centering 
    \includegraphics[width=0.80\columnwidth]{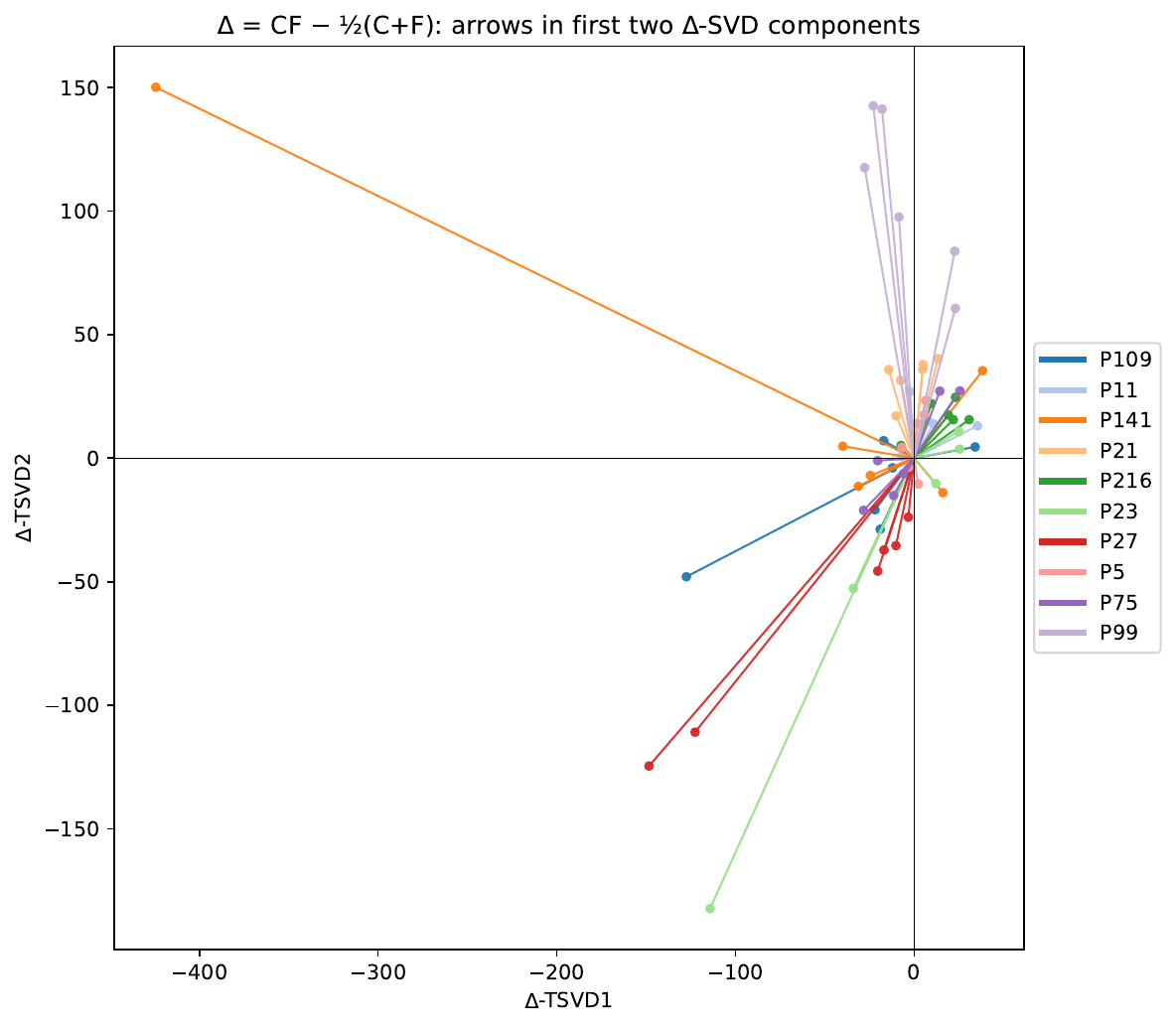} 
    \caption{Projection of patient-level $\Delta$ vectors into the first two singular directions. Each line segment extends from the origin to a patient replicate's $\Delta$ vector, colored by patient ID. The strong alignment along the negative $\Delta$-SVD1 axis highlights a coherent treatment-induced deviation shared across patients.} 
    \label{fig:delta_patient_proj} 
\end{figure}

Projecting patients into this $\Delta$-SVD space (Fig.~\ref{fig:delta_patient_proj}) reveals a striking asymmetry: most vectors align negatively along the first component ($\Delta$-SVD1), reflecting a shared population-level deviation under CF. Meanwhile, variation in lengths indicates heterogeneity across patients—some show strong aligned deviations, while others remain close to the origin, suggesting weaker departures from additivity.

\subsubsection{Bicluster Structure in the Top $\Delta$ Component}
\begin{figure}[ht]
    \centering
    \begin{minipage}{0.48\columnwidth}
        \centering
        \includegraphics[width=\linewidth]{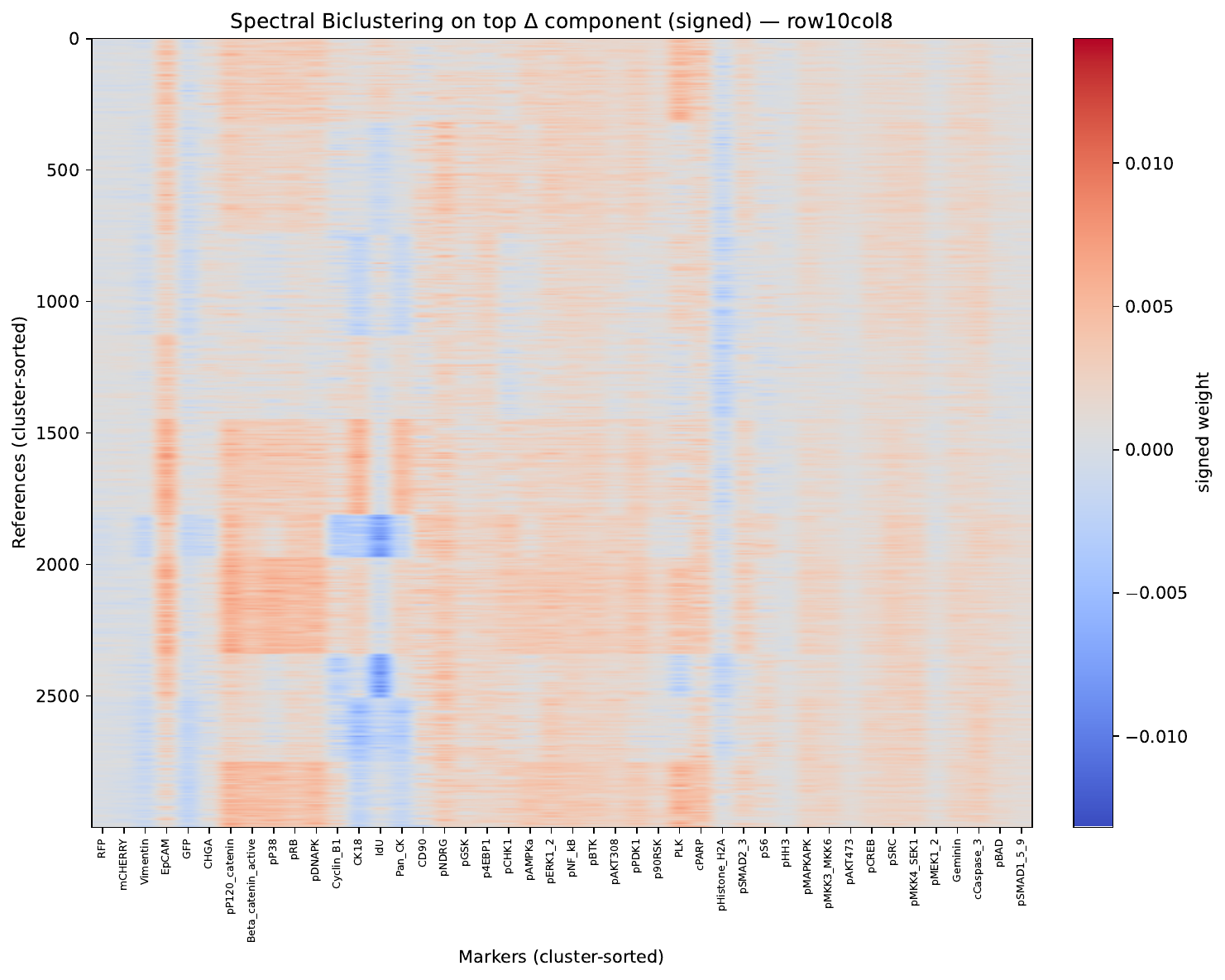}
        \caption*{\footnotesize (a) 10 row × 8 column clusters}
    \end{minipage}\hfill
    \begin{minipage}{0.48\columnwidth}
        \centering
        \includegraphics[width=\linewidth]{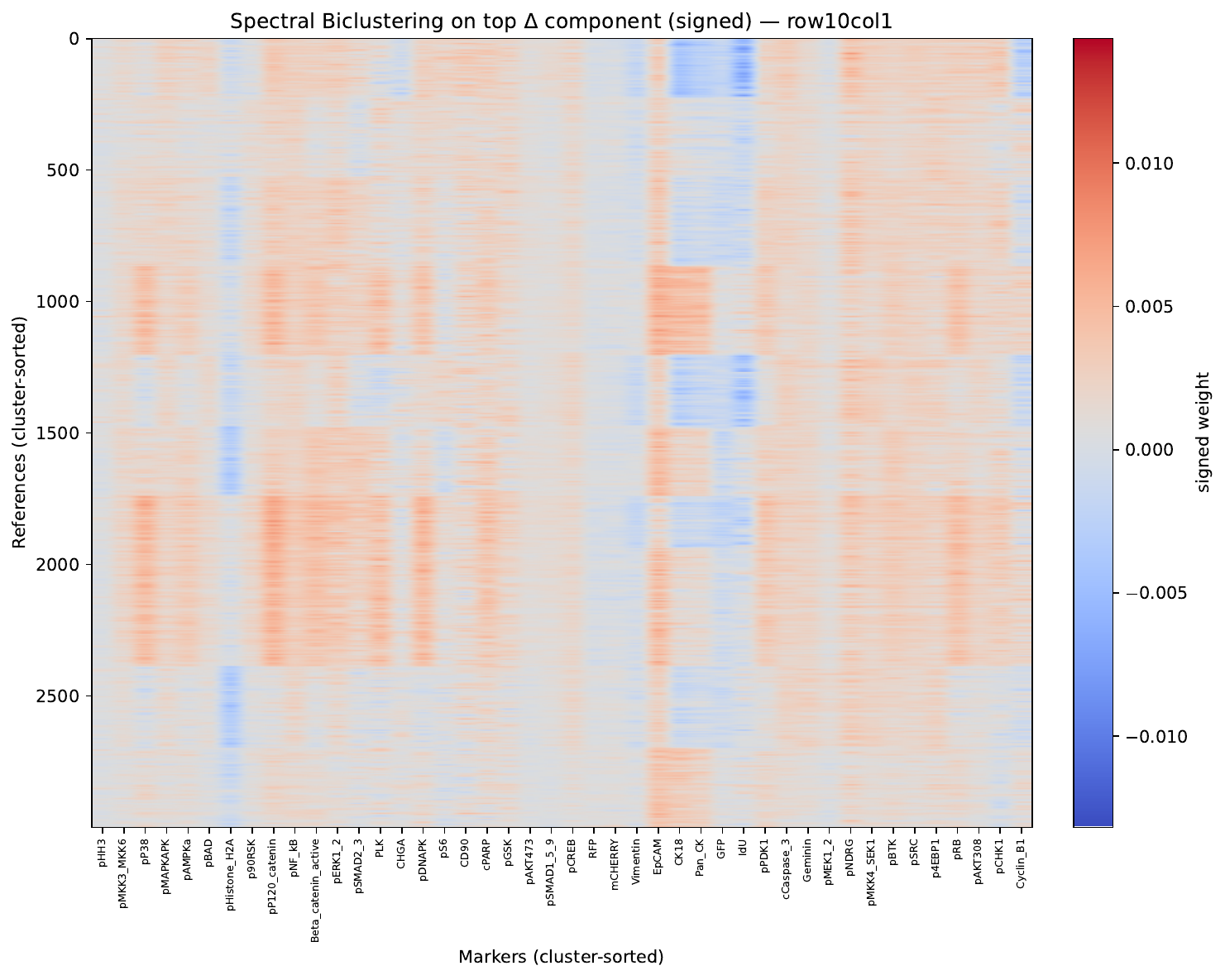}
        \caption*{\footnotesize (b) 10 row × 1 column cluster}
    \end{minipage}
    \caption{Spectral biclustering of the top $\Delta$ component (signed weights). 
    Dark red and blue blocks reveal subsets of markers and reference points with aligned positive or negative contributions, 
    highlighting structured deviations in the dominant $\Delta$-SVD1 direction.}
    \label{fig:pdo_bicluster}
\end{figure}

Spectral biclustering of the top $\Delta$ component revealed coherent blocks of reference–marker pairs with aligned signed weights (Figure~\ref{fig:pdo_bicluster}). 
The heatmaps show alternating red and blue patches, indicating groups of markers that jointly contribute positive or negative deviations in specific spatial regions. 
Dark red regions correspond to subsets of markers and references that consistently load positively on the $\Delta$-SVD1 axis, while dark blue regions highlight negatively weighted marker–reference combinations. The presence of these concentrated blocks suggests that the dominant treatment deviation is not diffuse across all features, but is instead driven by structured subsets of markers and spatial references. This structured organization supports the interpretation that the CF treatment engages specific signaling programs rather than producing uniform shifts across all features.

\section{Conclusion}

In this work, we demonstrated how Linear Optimal Transport (LOT) embeddings provide a powerful and interpretable framework for analyzing high-dimensional immune data.  The key strength of LOT is that it transforms irregular point clouds, which vary in size, shape, and lack alignment across samples, into a common Euclidean representation, making standard linear models directly applicable while still preserving meaningful distributional geometry.  
 
On COVID-19 immune profiling, LOT embeddings supported highly accurate classification of patient status. More importantly, because the classifier weights can be mapped back to reference points and biological markers, the discriminative power of the model can be traced to interpretable spatial–feature patterns.
This contrasts with many nonlinear methods (e.g., kernel machines, neural networks) that, while predictive, act as black boxes and cannot be directly connected to the original cellular data.  
Our pipeline shows that predictive performance and interpretability need not be mutually exclusive: LOT enables transparent classification where decisions can be explained in terms of marker-level differences across patient groups.

On patient-derived organoids (PDOs), LOT further enabled the construction of synthetic embeddings that test hypotheses about treatment effects.  
Although the original point clouds differ in size and structure, the uniform LOT representation allows averaging and interpolation to create new synthetic conditions.  
This offers a systematic approach to investigate beyond additive models and to discern organized, population-level variations caused by drug combinations.

Taken together, these findings show that LOT offers a coherent, interpretable, and generative framework for single-cell analysis, in addition to a predictive embedding. It bridges the gap between high accuracy and biological insight, enabling both classification of patient states and the creation of synthetic conditions, while remaining directly anchored to the original point-cloud data.  
We believe this approach can generalize to a wide range of biological and scientific domains where data are inherently distributional and interpretability is essential.

\bibliographystyle{plainnat}
\bibliography{references}

\end{document}